\documentclass{IOS-Book-Article}

\usepackage{mathptmx}
\usepackage{soul}\setuldepth{article}
\usepackage{graphicx}
\usepackage{todonotes}
\usepackage{soul,color}
\usepackage[section]{placeins}
\usepackage{multirow}
\usepackage{bm}
\usepackage{url}
\usepackage{siunitx}
\usepackage{tikz,forest}
\usepackage{graphicx}
\usepackage{times}
\usepackage{url}
\usepackage{latexsym}
\usepackage{url}
\usepackage{hyperref}
\usepackage{booktabs}
\usepackage{multirow}
\usepackage[T1]{fontenc}%
\usepackage[utf8]{inputenc}
\def\hb{\hbox to 10.7 cm{}}

\begin{document}

\pagestyle{headings}
\def\thepage{}

\begin{frontmatter}              

\newcommand{\XXX}[1]{\textcolor{red}{XXX #1}}

\title{Robust Neural Machine Translation: Modeling Orthographic and Interpunctual Variation}

\markboth{}{July 2020\hb}

\author[A,B]{\fnms{Toms} \snm{Bergmanis}%
\thanks{Corresponding Author: Toms Bergmanis; E-mail: toms.bergmanis@tilde.lv.}},
\author[A]{\fnms{Artūrs} \snm{Stafanovičs}},
\author[A,B]{\fnms{Mārcis} \snm{Pinnis}}

\runningauthor{Bergmanis et al.}
\address[A]{Tilde, Riga, Latvia}
\address[B]{Faculty of Computing, University of Latvia}

\begin{abstract}
Neural machine translation systems typically are trained on curated corpora and break when faced with non-standard orthography or punctuation. Resilience to spelling mistakes and typos, however, is crucial as machine translation systems are used to translate texts of informal origins, such as chat conversations, social media posts and web pages. We propose a simple generative noise model to generate adversarial examples of ten different types. We use these to augment machine translation systems' training data and show that, when tested on noisy data, systems trained using adversarial examples perform almost as well as when translating clean data, while baseline systems' performance drops by 2-3 BLEU points. To measure the robustness and noise invariance of machine translation systems' outputs, we use the average translation edit rate between the translation of the original sentence and its noised variants. Using this measure, we show that systems trained on adversarial examples on average yield 50\% consistency improvements when compared to baselines trained on clean data.
\end{abstract}
\end{frontmatter}
\markboth{July 2020\hb}{July 2020\hb}
\section{Introduction}
Humans exhibit resilience to orthographic variation in written text \cite{rawlinson1976significance,mccusker1981word}. 
As a result, spelling mistakes and typos are often left unnoticed. 
This flexibility of ours, however, is shown to be detrimental for neural machine translation (NMT) systems, which typically are trained on curated corpora and tend to break when faced with noisy data \cite{belinkov2018synthetic,michel2018mtnt}. Achieving NMT robustness to human blunder, however, is important when translating texts of less formal origins, such as chat conversations, social media posts and web pages with comment sections.

In this work, we propose, to augment NMT system's training data with data where source sentences are corrupted with adversarial examples of different types. There have been various studies on the impact of different types and sources of noise on NMT \cite{carpuat-etal-2017-detecting,khayrallah2018impact,zhou2019improving}. In this work, we focus on the noise caused by orthographic variation of words, such as unintentional misspellings and deliberate spelling alternations as well as noise due to misplaced and omitted punctuation. Thus, the closest to this study is the work on black-box adversarial training of NMT systems \cite{belinkov2018synthetic,sperber2017toward,vaibhav2019improving}, where models are trained on adversarial examples that are generated without accessing the model's parameters. 
Unlike the previous work, which focuses only on adversarial examples that model unintentional changes of spelling, we also model deliberate orthographic alternation, such as omission and substitution of diacritical signs. As we show in our experiments, such orthographic variation has a more substantial negative impact on MT outputs than the other types of noise and thus is more important to be accounted for. 
Further, to overcome the lack of curated evaluation datasets as required by the previous work \cite{michel2018mtnt,vaibhav2019improving}, we propose an automatic evaluation method that measures the noise invariance of MT outputs without relying on a reference translation. By measuring noise invariance of MT outputs the method also allows us to assess whether MT system translation consistency improves when facing small variations in the source text.

\begin{table}[h]
\caption{Noise applied to the example sentence: ``\textit{Balta jūra, zaļa zeme.}'' Were possible, noise is marked in bold, otherwise it is indicated with `\_'.}
\centering
\begin{tabular}{lll}
\toprule
\textbf{\#} & \textbf{Type}                    & \multicolumn{1}{c}{\textbf{Examples}} \\

\midrule
1  & introduce extra letters & Bal\textbf{z}ta jūra, zaļa zeme.      \\
2  & delete letters          & \textbf{\_}alta jūra, zaļa zeme.        \\
3  & permute letters         & Ba\textbf{tl}a jūra, zaļa zeme.       \\
4  & confuse letters         & Balta jūra, \textbf{x}aļa zeme.       \\
5  & add diacritic           & Balta jūra, zaļa z\textbf{ē}me.       \\
6  & sample substitute       & Balta jūra, zaļa \textbf{zemi}.       \\
\midrule
7 & remove punctuation      & Balta jūra\textbf{\_} zaļa zeme\textbf{\_}         \\
8 & add comma               & Balta\textbf{,} jūra, zaļa zeme.     \\
\midrule
9  & latinize                & Balta j\textbf{u}ra, za\textbf{l}a zeme.       \\
10 & phonetic latinize       & Balta j\textbf{uu}ra, za\textbf{lj}a zeme.     \\ 
\bottomrule
\end{tabular}

\label{table:noise_examples}
\end{table}

\section{Methods} \label{section:methods}
We propose a simple generative noise model to generate adversarial examples of ten different types. These include incidental insertion, deletion, permutation and keyboard-based confusion of letters as well as the addition of a diacritic to letters which support them (Table~\ref{table:noise_examples}, examples 1-5). We also explicitly model the misspellings that result in another valid word (Table~\ref{table:noise_examples}, example 6). For interpunctual variation, we consider sentences with missing punctuation and incorrectly placed commas (Table~\ref{table:noise_examples}, examples 7-8). For deliberate orthographic changes, we support sentence-level omission and phonetic latinization of diacritical signs (Table~\ref{table:noise_examples}, examples 9-10). 

\paragraph{Measure of Robustness} To measure NMT robustness and noise invariance of NMT outputs, we calculate the average translation edit rate (TER) \cite{snover2006study} between the translation of the original orthographically correct sentence and the translations of its ten noised variants for each noise type. We refer to it as tenfold noisy translation TER, or \textbf{10NT-TER}. This measure gives a score of 0 if all ten translations of a sentence with added noise match the translation of the original sentence and a score of 100 (or more) if all of them had no word in common with the translation of the original sentence.

\section{Experimental Setting} 
\begin{table}[t]
\caption{The original training data sizes and data sizes with adversarial examples included.}
\centering
\begin{tabular}{llcc}
\toprule
                            &                    & \multicolumn{2}{c}{\textbf{Train}} \\
                            &                    & Original  & \multicolumn{1}{l}{+ adversarial noise} \\
\midrule
Small data                  & English-Latvian    & 4.5M      & 9M                                      \\
\midrule

\multirow{3}{*}{Large data} & English-Estonian   & 34.9M     & 69.8M                                   \\
                            & English-Latvian    & 45.2M     & 90.4M                                   \\
                            & English-Lithuanian & 22.1M     & 44.2M                                  \\
                            \midrule
\end{tabular}

\label{tab:data}
\end{table}

\paragraph{Languages and Data}
We conduct experiments on Estonian-English, Latvian-English and Lithuania-English language pairs. 
We use the Latvian-English constrained data from the WMT~2017\footnote{\url{http://www.statmt.org/wmt17}} news translation shared task to train \textbf{small data systems} that we use for development and analysis of our methods. To verify that our findings also hold not only for small data settings, but also production-grade systems that are trained on much larger data, we use large datasets from the Tilde Data Library\footnote{\url{https://www.tilde.com/products-and-services/data-library}} to train \textbf{large data systems}. 
For validation during training and testing, we use development and test sets from the WMT news translation shared tasks. For English-Estonian, we use the data from WMT~2018, for English-Latvian -- WMT~2017, and for English-Lithuanian -- WMT~2019\footnote{\url{http://www.statmt.org/wmt17|18|19}}.

We use a simplified and production-grade data pre-processing pipelines. The simplified data pre-processing consists of the standard Moses \cite{koehn2007moses} scripts for tokenization, cleaning, normalization, and truecasing, while the production grade pipeline consists of Tilde MT platform's \cite{pinnis2018tilde} implementation of the same processes.

\paragraph{NMT Models} We mostly use the default configuration\footnote{\url{https://github.com/marian-nmt/marian-examples/tree/master/transformer}} of the Marian \cite{junczys2018marian} toolkit’s  implementation of the Transformer model \cite{vaswani2017attention}. We select batch sizes dynamically so that they fit in a workspace of 9000MB. Additionally, we use delayed gradient updates \cite{bogoychev-etal-2018-accelerating} by setting optimizer delay to 4. We stop  model training after ten consecutive evaluations with no improvement in translation quality on the development set \cite{prechelt1998early}. 

\section{Experiments}
\paragraph{Initial Experiments}
To test the effect of individual noise models on MT systems' performance, we train separate Latvian-English small data systems on original data augmented in a 1-to-1 proportion with each type of adversarial examples. All in all, we obtain ten systems trained using adversarial examples and the baseline. We test each system on the original development set and development sets that have adversarial examples of each type of noise. Table~\ref{tab:lv-all-noise} summarises the results. First, we note that including adversarial examples improves the overall translation quality and especially quality on development sets containing the adversarial examples that the systems have seen during training.

\begin{table}[]
\caption{Latvian-English development set results in BLEU \cite{papineni2002bleu} points for small data systems. Rows: systems trained on original data that are 1:1 up-sampled with each type of adversarial examples. Columns: development sets with each type of adversarial examples.}
\label{tab:lv-all-noise}
\centering
\includegraphics[width=1\linewidth]{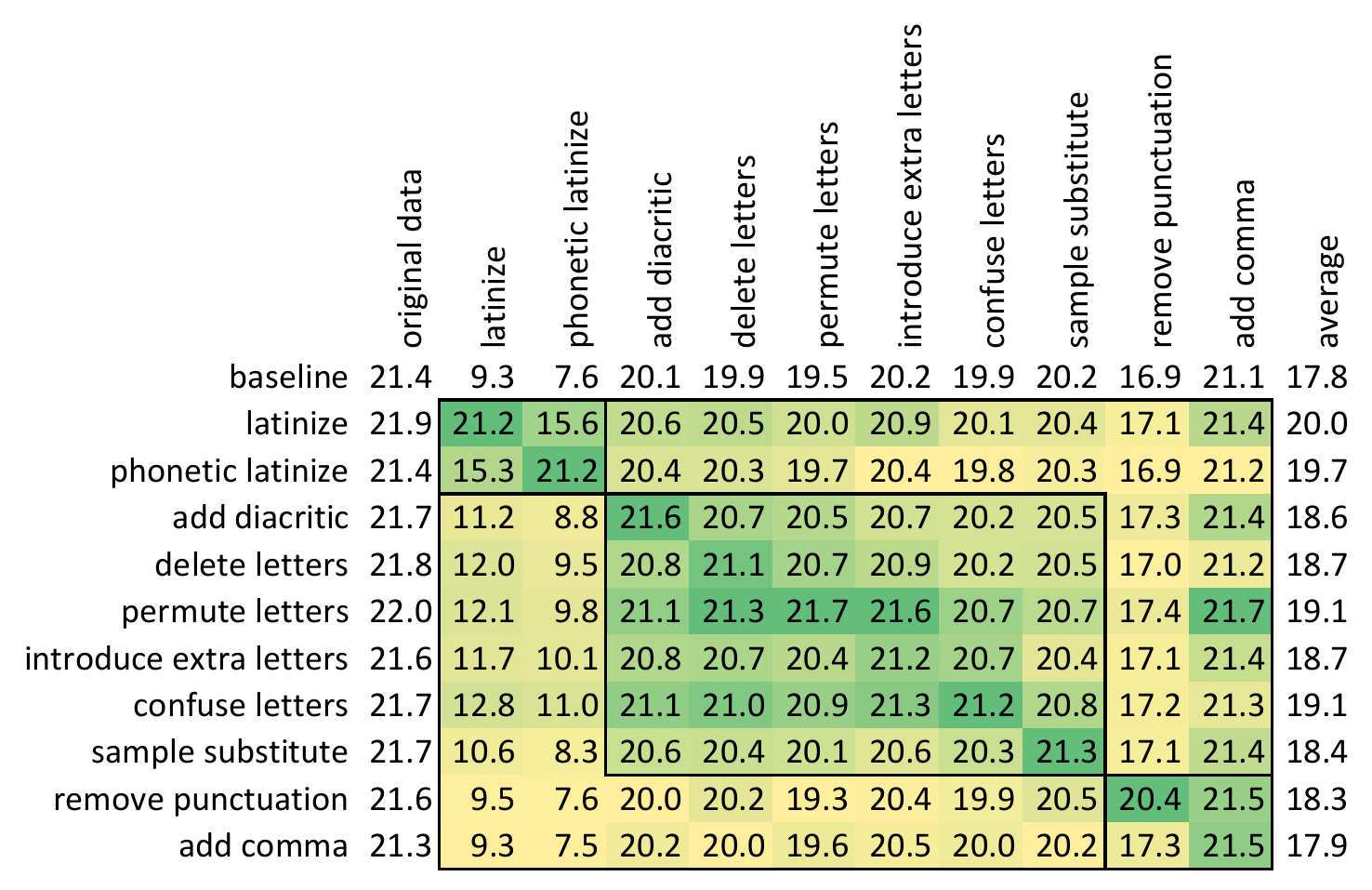}
\end{table}

Second, we observe that not all diagonal elements of Table~\ref{tab:lv-all-noise} contain the highest BLEU score for their respective column, suggesting existing redundancies between the noise models.  Examples are MT systems trained using adversarial examples from noise models that \textit{delete letters} and \textit{introduce extra letters}, which both when tested on their respective adversarial example development sets come second to the MT system that was trained using adversarial examples from the noise model that \textit{permutes letters} (21.1 vs 21.3~BLEU points and 20.9 vs 21.6~BLEU points respectively). Similarly, the MT system trained using the noise model that \textit{adds a comma} (21.5~BLEU), shows no benefit over the system that was trained using examples from the model that \textit{removes punctuation} (21.5~BLEU). Based on these results, we decided not to include the redundant models (\textit{delete letters}, \textit{introduce extra letters} and \textit{add comma}) in further experiments. 

We, however, also recognize that the performance gains caused by the remaining noise models are numerically small (+0.5~BLEU) when compared against the next best performing MT system. For this reason, we use bootstrap re-sampling \cite{koehn2004statistical} to test if the performance gains of MT systems trained on adversarial examples generated by the noise models that \textit{add a diacritic}, \textit{confuse letters}, and perform \textit{sample substitution} are statistically significant if compared against a system that is trained on adversarial examples generated by the noise model that \textit{permutes letters}. Tests confirm that all differences are indeed significant at $p<0.05$. Based on these tests, we include these models in our final experiments.

\begin{table}[]
\caption{Test set results in BLEU points for large data MT systems.}
\label{tab:bleu}
\centering
\includegraphics[width=1\linewidth]{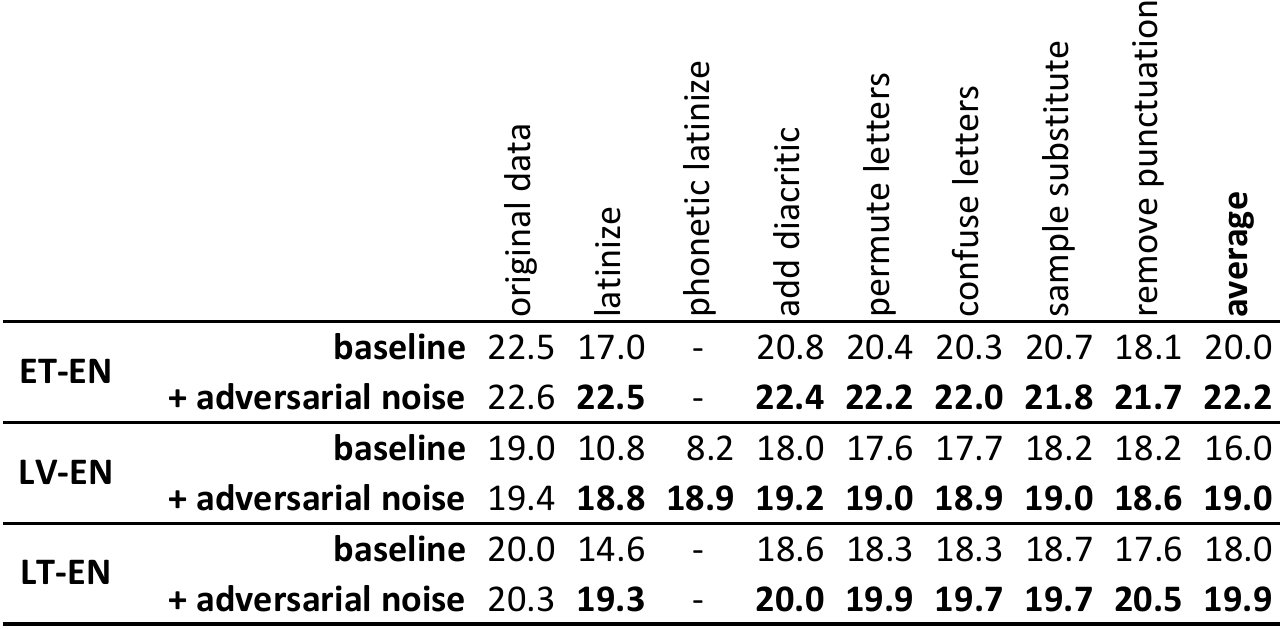}

\end{table}
\paragraph{Large Data Systems}
To test the effect of the seven productive noise models on MT system translation quality, we train Estonian-English, Latvian-English and Lithuanian-English large data MT systems. For systems trained using adversarial examples, we augment the original data with another copy of the data in which each type of noise is applied at an equal proportion. The evaluation results of these systems on the test data sets are provided in Figure~\ref{tab:bleu}. First, we observe that the performance of the baseline MT systems and systems trained using adversarial examples on the original test data sets is comparable, suggesting that using adversarial examples does not harm the translation of clean data. 
Next, we observe, that when tested on noisy data systems that are trained using adversarial examples perform only slightly worse (about -0.4 BLEU on average) than when translating clean data, while baseline systems show an average performance drop of about 2 BLEU points. 

\begin{table}[]
\centering
\caption{Examples of noise in Latvian language input data causing widely different English language translations.}
\label{tab:examples}
\begin{tabular}{ll}
\toprule
\textbf{Orig.} & Twitter lietotāji nespēja noticēt, dzirdot Bairona Makdonalda neiejūtīgos komentārus.         \\
\textbf{Ref.} &  Twitter users did not hold back when they heard how insensitive Byron Macdonlad was being. \\
\midrule
\textbf{Src. }   & Twitter \textbf{leitotāji} nespēja noticēt, dzirdot Bairona Makdonalda neiejūtīgos komentārus. \\
\textbf{Hyp.}   & Twitter's lieutenants couldn't believe it by hearing Byron McDonald's insensitive comments.   \\
\midrule
\textbf{Src.}    & Twitter lietotāji nespēja noticēt, \textbf{dzidrot} Bairona Makdonalda neiejūtīgos komentārus. \\
\textbf{Hyp. }   & Twitter users could not believe by clarifying Byron McDonald's insensitive comments.  \\
\midrule
\textbf{Src.}    & \textbf{Twtiter} lietotāji nespēja noticēt, dzirdot Bairona Makdonalda neiejūtīgos komentārus. \\
\textbf{Hyp.}   & Twtiter users couldn't believe it when they heard Byron McDonald's insensitive comments.      \\
\midrule
\textbf{Src}.    & Twitter \textbf{liettoāji} nespēja noticēt, dzirdot Bairona Makdonalda neiejūtīgos komentārus. \\
\textbf{Hyp.}   & The Twitter countryside couldn't believe it by hearing Byron McDonald's insensitive comments. \\
\bottomrule
\end{tabular}

\end{table}

\paragraph{Robustness and Noise Invariance}
Besides measuring the changes in translation quality caused by noisy input data, we also would like to measure the robustness and noise invariance of the MT systems. We motivate this by the observation that often small perturbations in input data lead to widely different MT outputs (see Table~\ref{tab:examples}). Desideratum, however, is that MT system outputs are noise invariant to an extent at least that unintentional changes in input data do not affect the meaning of the translation output.
Results (see Table~\ref{tab:ter}) of our experiments show that using adversarial examples in training improves the robustness and noise invariance of the MT systems measured in 10NT-TER (see Section~\ref{section:methods}) on average by 0.1 10NT-TER points or in relative terms an average consistency improvement of about 50\%.

\begin{table}[h]
\caption{Robustness and noise invariance of large data MT systems measured in 10NT-TER (see Section~\ref{section:methods}). }
\label{tab:ter}
\centering
\includegraphics[width=1\linewidth]{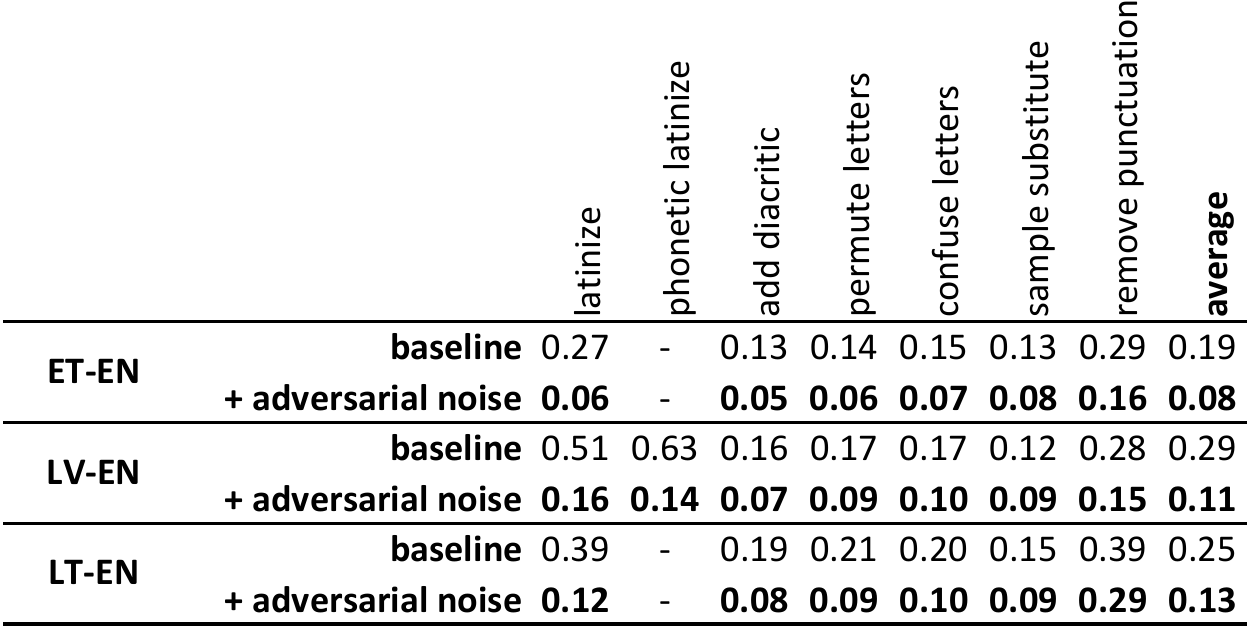}
\end{table}

\section{Conclusions}
We have proposed a simple generative noise model for the generation of adversarial examples for training data augmentation of NMT systems. 
Our results demonstrate that NMT systems that are trained using adversarial examples are more resilient to noisy input data. We show that while for the baseline NMT systems, noisy inputs cause a substantial drop in the translation quality (a drop of 2-3 BLEU points), for the systems that are trained using adversarial examples translation quality changes comparatively little (an average of -0.4 BLEU). In terms of translation robustness, systems trained on adversarial examples on average yield 50\% consistency improvement when compared to baselines trained on clean data. Methods proposed here will be useful for achieving NMT robustness to orthographic and interpunctual variation in input data. This will be especially beneficial in use cases where NMT systems are used to translate texts of informal origins, such as chat conversations, social media posts and web pages with comment sections.

\section{Acknowledgments}
This research has been supported by the European Regional Development Fund within the joint project of SIA TILDE and University of Latvia “Multilingual Artificial Intelligence Based Human Computer Interaction” No. 1.1.1.1/18/A/148.

\bibliographystyle{ios1}           
\bibliography{bibliography}

\end{document}